\documentclass{article}

\usepackage[preprint,nonatbib]{neurips_2021}
\usepackage{multirow}

\usepackage[utf8]{inputenc} 
\usepackage[T1]{fontenc}    
\usepackage{hyperref}       
\usepackage{url}            
\usepackage{booktabs}       
\usepackage{amsfonts}       
\usepackage{nicefrac}       
\usepackage{microtype}      
\usepackage{xcolor}         
\usepackage{graphicx}
\title{Learning from Synthetic Data: Facial Expression Classification based on Ensemble of Multi-task Networks}

\author{%
  Jae-Yeop Jeong${^1}$, Yeong-Gi Hong${^1}$, JiYeon Oh${^2}$, Sumin Hong${^3}$, and Jin-Woo Jeong${^1}$\\
  Department of Data Science${^1}$, Division of IISE${^2}$, Division of ITM${^3}$\\
  Seoul National University of Science and Technology\\
  Seoul, Korea \\
  \texttt{\{jaey.jeong, yghong, dhwldus0906, 17101992, jinw.jeong\}@seoultech.ac.kr} \\
   \And
   Yuchul Jung \\
   Department of Computer Engineering\\
   Kumoh National Institute of Technology \\
   Gumi, Korea \\
   \texttt{jyc@kumoh.ac.kr} \\
}

\begin{document}

\maketitle

\begin{abstract}
Facial expression in-the-wild is essential for various interactive computing domains. Especially, "Learning from Synthetic Data" (LSD) is an important topic in the facial expression recognition task. In this paper, we propose a multi-task learning-based facial expression recognition approach which consists of emotion and appearance learning branches that can share all face information, and present preliminary results for the LSD challenge introduced in the 4th affective behavior analysis in-the-wild (ABAW) competition. Our method achieved the mean F1 score of 0.71.
  
\end{abstract}

\section{Introduction}
Facial expression recognition plays an important role in various interactive computing domains such as human-computer interaction, social robots, and a satisfaction survey \cite{li2020deep}. To achieve more robust and accurate facial expression recognition (FER), a number of studies have been proposed in recent years \cite{savchenko2022classifying, wen2021distract, zhou2019exploring, kollias2021distribution, kollias2021affect, kollias2017recognition}. However, there are still many rooms to improve the robustness and performance of FER techniques. One of the most challenging research areas is facial expression recognition in-the-wild. Basically, to get high-level performance in FER, a number of well-aligned and high-resolution face images are necessary. Compared to face images that are gathered in a controlled setting, however, in-the-wild face images have many variations, such as various head poses, illumination, etc. Therefore, facial expression recognition in-the-wild is still challenging and should be studied continuously for real-world applications. Accordingly, face image generation/synthesis for FER tasks has been steadily getting much attention because it can generate unlimited photo-realistic facial images with various expressions under various conditions \cite{abbasnejad2017using, Zeng_2018_ECCV, gao2022face}. By learning from synthetic data and evaluating on the real-world data, the problem of construction of large-scale real data sets would be mitigated.

The 4th competition on Affective Behavior Analysis in-the-wild (ABAW), held in conjunction with the European Conference on Computer Vision (ECCV) 2022 \cite{kollias2022abaw_eccv}, a continuation of 3rd Workshop and Competition on Affective Behavior Analysis in-the-wild (ABAW) in CVPR 2022 \cite{kollias2022abaw}. The ABAW competition contributes to the deployment of in-the-wild affective behavior analysis systems that are robust to video recording conditions, diversity of contexts and timing of display, regardless of human age, gender, ethnicity, and status. The 4th ABAW competition is based on the Aff-Wild2 database \cite{kollias2019expression}, which is an extension of the Aff-wild database \cite{zafeiriou2017aff, kollias2019deep} and consists of the following tracks: 1) Multi-Task-Learning (MTL) 2) Learning from Synthetic Data (LSD). 

In this paper, we describe our methods for the LSD challenge and present preliminary results. For the LSD challenge, some frames from the Aff-Wild2 database \cite{kollias2019expression} were selected by the competition organizers and then used to generate artificial face images with various facial expressions \cite{kollias2020deep, kollias2020va, kollias2018photorealistic}. In total, the synthetic image set consists of approximately 300K images and their corresponding annotations for 6 basic facial expressions (anger, disgust, fear, happiness, sadness, surprise), which will be used in model training/methodology development. 
In this year's LSD challenge, participating teams were allowed to use only the provided synthetic facial images when developing their methodology, while any kind of pre-trained model could be used unless it was not been trained on the Aff-Wild2 database. For validation, a set of original facial images of the subjects who also appeared in the training set was provided. For evaluation, the original facial images of the subjects in the Aff-Wild2 database test set, who did not appear in the given training set, are used. For the LSD challenge, the mean F1 score across all 6 categories was used as metric.

\section{Method}
\subsection{Overview}
\begin{figure}[t]
    \centering
    \includegraphics[width=\columnwidth]{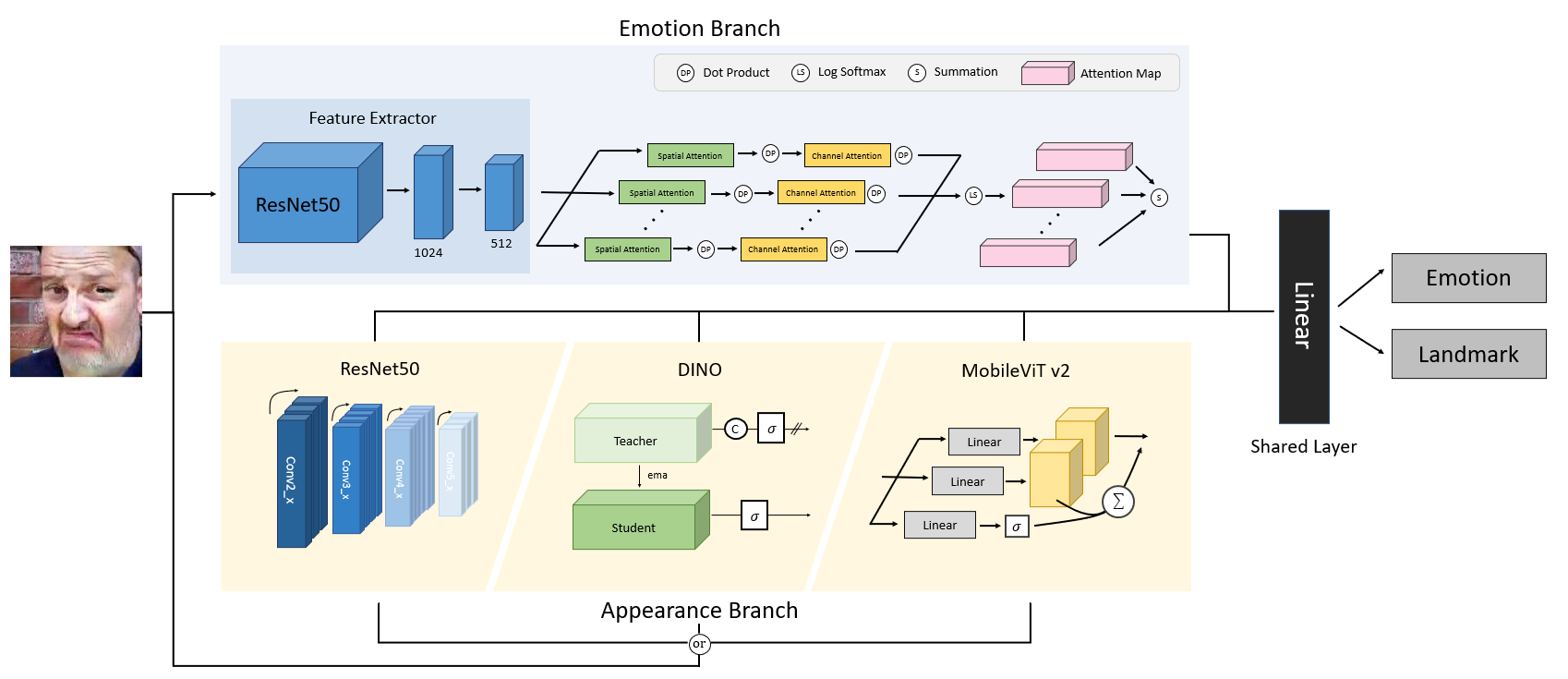}
    \caption{Overview of the architecture used in this study}
    \label{fig:architecture}
\end{figure}

\begin{figure}[t]
    \centering
    \includegraphics[width=\columnwidth]{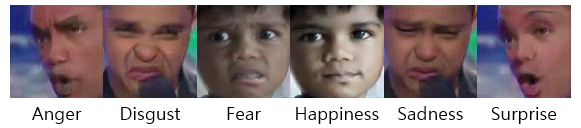}
    \caption{Synthesis data used in our study}
    \label{fig:dataset}
\end{figure}

To achieve a high performance on the task of facial expression recognition, extraction of the robust feature from input facial images is essential. To this end, we employed a multi-task learning approach, jointly optimizing different learning objectives. Figure \ref{fig:architecture} depicts an overview of the proposed architecture used in our study. 
As shown in Figure \ref{fig:architecture}, the framework is trained to solve both facial expression recognition task (i.e., Emotion branch) and face landmark detection task (i.e., Appearance branch) using the provided synthetic training data only. Figure \ref{fig:dataset} shows the sample provided training images.
Our multi-task learning framework has two output layers for tasks of FER and landmark detection, $y_{expr}$ and $y_{land}$ respectively. Here, $y_{expr} \in  R^{6}$ and $y_{land} \in  R^{68 \times 2}$. Finally, the loss of $y_{expr}$ and $y_{land}$ are computed with weighted cross-entropy and mean squared error, respectively.
During the inference, only the output from the emotion branch is used for classification of facial expression for the given validation/test sample.

To achieve a more generalized performance for in-the-wild data, we applied a bagging approach when generating final predictions.
Each model with a different configuration was trained with sub-sampled data sets (i.e., 20\% out of the entire set) and produced its own output. Finally, we aggregate the probabilities of each model through soft voting for the final prediction.
More details on the CNN models used in each branch can be found from Section \ref{modelarch}.
  

\subsection{Database}
\begin{figure}[t]
    \centering
    \includegraphics[scale=0.5]{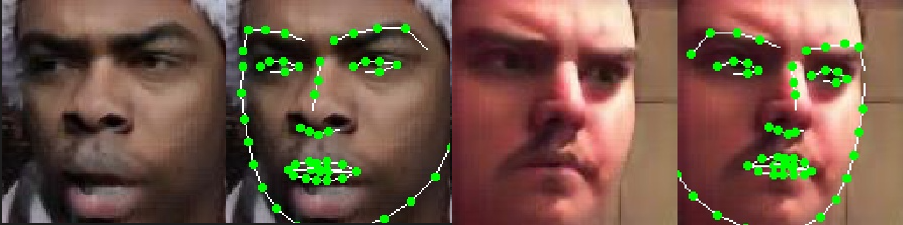}
    \caption{Example of face landmarks}
    \label{fig:landmark}
\end{figure}

In order to train our multi-task learning framework, each image is given 1) facial expression annotation and 2) face landmark annotation. First, the affective state label consists of 6 basic facial expressions (anger, disgust, fear, happiness, sadness, surprise), which were offered by the 4th ABAW competition organizers. Next, the landmark annotations were created through DECA \cite{feng2021learning} framework which is a state-of-the-art 3D face shape reconstruction method that reconstructs both face landmark and 3D mesh from a single facial image. Figure \ref{fig:landmark} shows our example of landmark annotation which is composed of 68 coordinate points with (x, y) of the face.

\subsection{Model Architecture}
\label{modelarch}
As depicted in Figure \ref{fig:architecture}, our architecture is composed of two branches: 1) emotion and 2) appearance. In each branch, we utilized a pre-trained backbone for a robust and generalized feature extraction. Finally, we employed a series of shared fully connected layers right after two branches, to exploit all knowledge extracted from the backbone model of each branch.

\subsubsection{Emotion branch}
\label{sec:emtion_branch}
As depicted in Figure \ref{fig:architecture} emotion branch, we adopted a deep learning-based FER approach called "DAN" \cite{wen2021distract} which is a state-of-the-art method for the AffectNet database \cite{mollahosseini2017affectnet}. The DAN architecture has two phases: feature extractor and attention parts. In the attention phase, there exist multi-head cross attention units which consist of a combination of spatial and channel attention units. The DAN architecture used in our emotion branch is a modified version pre-trained on AffectNet \cite{mollahosseini2017affectnet}, Expw \cite{expw}, and AIHUB datasets \cite{aihub} for a categorical FER task. For better performance, we replaced the original feature extractor of the DAN architecture (i.e., ResNet18 \cite{he2016deep} pre-trained on MS-Celeb-1M \cite{guo2016ms}) with ResNet50 \cite{he2016deep} pre-trained on VGGFace2 \cite{cao2018vggface2}, as presented in \cite{jeong2022classification}.
To prevent overfitting and acquire a generalizable performance, we applied various data augmentation techniques, such as Colorjitter, Horizontal flip, RandomErasing, and Mix-Augment \cite{Psaroudakis_2022_CVPR}. Finally, the input data go through our deep network in the emotion branch and are fed to the shared fully connected layers.

\subsubsection{Appearance branch}
The goal of the appearance branch is to extract the robust feature in terms of visual appearance. For this, we employed various backbone models pre-trained on large scale data sets, ResNet50 \cite{he2016deep} pre-trained on VGGFace2 \cite{cao2018vggface2}, DINO ResNet50 \cite{caron2021emerging} pre-trained on VGGFace \cite{parkhi2015deep}, and MobileVITv2 \cite{mehta2022separable} pre-trained on ImageNet \cite{krizhevsky2012imagenet}, as a feature extractor for the appearance branch. 
As shown in the appearance branch of Figure \ref{fig:architecture}, we trained only a single backbone for landmark detection during the multi-task learning phase, rather than utilizing multiple backbones together.
Similar to the emotion branch, we also used a data augmentation strategy in the appearance branch. Due to the characteristics of landmark data, we applied Colorjitter only to prevent unnecessary spatial transformations. After feature extraction, all the landmark-related features are passed to the shared fully connected layers. 

In summary, our multi-task learning model is configured with DAN on the emotion branch and the appearance branch with one of the following backbones: a) ResNet50, b) DINO ResNet50, c) MobileViT-v2.

\section{Results}
All the experiments were conducted using a GPU server with six NVIDIA RTX 3090 GPUs, 128 GB RAM, Intel i9-10940X CPU, and Pytorch framework. Our preliminary results on the official validation set for the LSD challenge was 0.71 in terms of the mean F1 score, which significantly outperforms that of a baseline method (0.50). 

%


\section{Conclusion}

In this paper, we proposed a multi-task learning-based architecture for FER and presented the preliminary results for the LSD challenge in the 4th ABAW competition. Our method produced the mean F1 score of 0.71 for the LSD challenge. The implementation details and validation results may be updated after submission of this paper to arxiv.

\nocite{*}
{\small
\bibliographystyle{ieee_fullname}
\bibliography{egbib}
}
\end{document}